\documentclass[letterpaper]{article} % DO NOT CHANGE THIS
\usepackage{aaai25}  % DO NOT CHANGE THIS
\usepackage{times}  % DO NOT CHANGE THIS
\usepackage{helvet}  % DO NOT CHANGE THIS
\usepackage{courier}  % DO NOT CHANGE THIS
\usepackage[hyphens]{url}  % DO NOT CHANGE THIS
\usepackage{graphicx} % DO NOT CHANGE THIS
\usepackage{color}
\usepackage{natbib}  % DO NOT CHANGE THIS AND DO NOT ADD ANY OPTIONS TO IT
\usepackage{caption} % DO NOT CHANGE THIS AND DO NOT ADD ANY OPTIONS TO IT
\usepackage{algorithm}
\usepackage{algorithmic}

\usepackage{newfloat}
\usepackage{listings}

\DeclareCaptionStyle{ruled}
{labelfont=normalfont,labelsep=colon,strut=off} % DO NOT CHANGE THIS
\floatstyle{ruled}
\newfloat{listing}{tb}{lst}{}
\floatname{listing}{Listing}
\usepackage{xcolor}

\title{ An LLM-Guided Tutoring System for Social Skills Training }
\author {
     Michael Guevarra\textsuperscript{\rm 1},
     Indronil Bhattacharjee\textsuperscript{\rm 2},
     Srijita Das\textsuperscript{\rm 3},
     Christabel Wayllace\textsuperscript{\rm 2},\\
     Carrie Demmans Epp\textsuperscript{\rm 4},
     Matthew E.~Taylor\textsuperscript{\rm 4,5},
     Alan Tay\textsuperscript{\rm 1}
 }
 \affiliations {
     \textsuperscript{\rm 1}  Illumia Labs,  \\
     \textsuperscript{\rm 2} New Mexico State University, \\
     \textsuperscript{\rm 3} University of Michigan-Dearborn, \\
     \textsuperscript{\rm 4} University of Alberta, \\
     \textsuperscript{\rm 5} Alberta Machine Intelligence Institute (Amii) \\
     \{michael.gue, alan.tay\}@illumialabs.ai, \{indronil, cwayllac\}@nmsu.edu, sridas@umich.edu, \{demmanse, matthew.e.taylor\}@ualberta.ca
}
\usepackage{bibentry}
\begin{document}

\maketitle

\begin{abstract}
Social skills training targets behaviors necessary for success in social interactions. However, traditional classroom training for such skills is often insufficient to teach effective communication --- one-to-one interaction in real-world scenarios is preferred to lecture-style information delivery. This paper introduces a framework that allows instructors to collaborate with large language models to dynamically design realistic scenarios for students to communicate. Our framework uses these scenarios to enable student rehearsal, provide immediate feedback, and visualize performance for both students and instructors. Unlike traditional intelligent tutoring systems, instructors can easily co-create scenarios with a large language model without technical skills. Additionally, the system generates new scenario branches in real time when existing options do not fit the student's response. 
% Our framework then uses these scenarios, enabling student rehearsal, providing immediate student evaluation and feedback, and visualizing performance for the instructor or student. In contrast to traditional intelligent tutoring systems, we enable instructors to easily co-create scenarios with a large language model without programming expertise. Furthermore, the system can generate new scenario branches in real time when existing options do not apply to an individual student's response.
\end{abstract}


\section{Introduction}
\noindent In today's rapidly evolving digital era, twenty-first century skills like communication and social skills have become critical~\cite{partner:19}, enhancing employee engagement and productivity~\cite{indeed:24,thornhill:23}. However, traditional classroom training for such soft skills is often insufficient to teach effective communication to deal with many real world scenarios. Such programs usually comprise a structured curriculum that often lacks the dynamics of the real-world environment within which these skills are effective and applicable.  AI-powered tools can bridge this gap by providing students with interactive, real-world scenarios, evaluate performance, and offer targeted feedback to further enhance their skills. Moreover, such AI tools can be effectively developed with an instructor-in-the-loop to help build or tailor domain-specific exercises.

Recent breakthroughs in large language models (LLMs) demonstrate their potential for conversational training and narrative generation. For instance, GENEVA~\cite{Leandro:24} uses LLMs to create graphs for branching narratives in dialogue-based role-playing games, allowing conversations to adapt dynamically within predefined constraints. Other work uses LLMs to simulate realistic agent behavior with complex social interactions in real time~\cite{Park:23}. Moreover,~\citet{Sun:24} suggests that personalizing conversational agents with individual personas enhances realism and engagement in training simulations. These applications show how LLMs can create adaptive and interactive environments for social skills training, providing realistic and context-specific scenarios for learners.

% \noindent Chris: One para on Related work goes in here.
Social skills training (SST)~\cite{tenhula:08} focuses on improving key behaviors necessary for success in social interactions. Behavioral SST methods include instruction, modeling, behavior rehearsal, feedback, and reinforcement, often combined with interpersonal problem-solving and social perception skills training. These skills are rarely taught using intelligent tutoring systems (ITSs), which provide immediate, customized instruction, or feedback. Previous work has used ITSs to teach counseling skills to U.S.\ officers~\cite{georgila:19}, assist children with social challenges~\cite{sanchez:14}, train social skills~\cite{tanaka:23}, and develop interpersonal and intercultural skills~\cite{lane:07}. 

This paper introduces a framework that leverages LLMs to train social skills. In contrast to traditional ITSs, our system empowers human instructors to dynamically design realistic situations, provides online scenarios for students to navigate, supports student rehearsal, offers immediate feedback, and includes a visualization tool for delayed feedback. Furthermore, unlike previous approaches, our system allows instructors to easily create and modify scenarios without programming expertise, providing the flexibility to tailor training to students' personalized needs and specific learning objectives.
%Writing the First para

% \noindnet Motivation (example)\\
% first para: Srijita
% Related work: Chris
% Related work\\
% Motivate from the perspective of building 21st century soft skills.
% Related work : Conversational agents, building softskill using AI, tutoring systems for building soft skills, common ication skill
% \textbf{@1st century softskills, LLM, pre-LLM, AIDE paper Chris}

 %System Flow Diagram\\
\begin{figure}[tbp]
    \centering
    \includegraphics[width=1\linewidth]{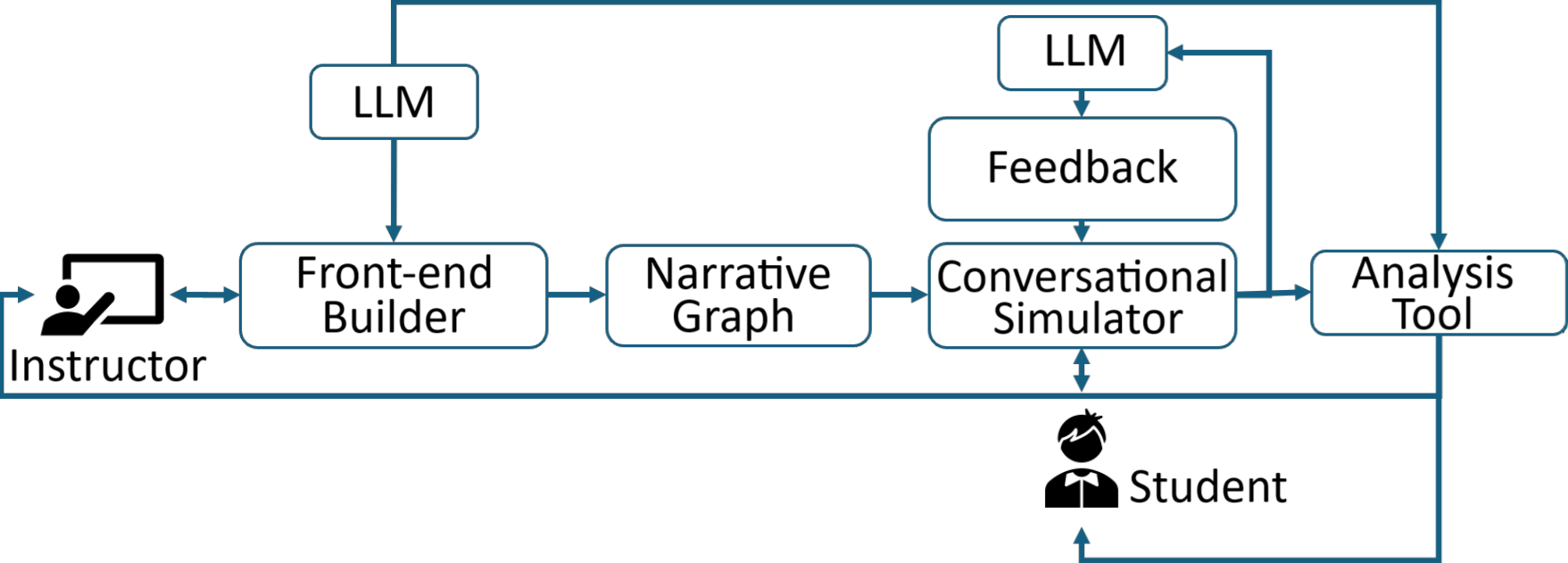}
    \caption{GLOSS: Block Diagram}
    \label{fig:blockDiag}
\end{figure}
\section {Guided Learning for Optimizing Social Skills}
Our proposed framework,  \underline{G}uided \underline{L}earning for \underline{O}ptimizing \underline{S}oft \underline{S}kills (GLOSS) is shown in  Figure~\ref{fig:blockDiag}. It consists of a front-end builder that allows an instructor and LLM to collaboratively build a narrative graph, which summarizes all possible interactions within the relevant scenario. When the student practices soft skills, a conversational simulator is used to show relevant dialogue from the scenario, either specified by the instructor or generated by the LLM along with appropriate feedback for the student's response. We implement various components of GLOSS using GPT-4.
%Components:
\\
\noindent \textbf{Front-end builder:} The front-end builder enables instructors to design scenarios that simulate students' interactions with people in a specific domain, such as handling an angry customer. Instructors can choose to use a pre-built template with scripted dialogue, create a scenario from scratch, generate a scenario using a prompt to an LLM, or combine these methods. Additionally, instructors can implement open-ended training scenarios, which offer higher training value but have traditionally been avoided due to the increased assessment load. This flexibility reduces content creation time, allowing for the development of more complex and realistic scenarios tailored to learning needs.
\begin{figure}[tbp]
    \centering
    \includegraphics[width=1.0\linewidth]{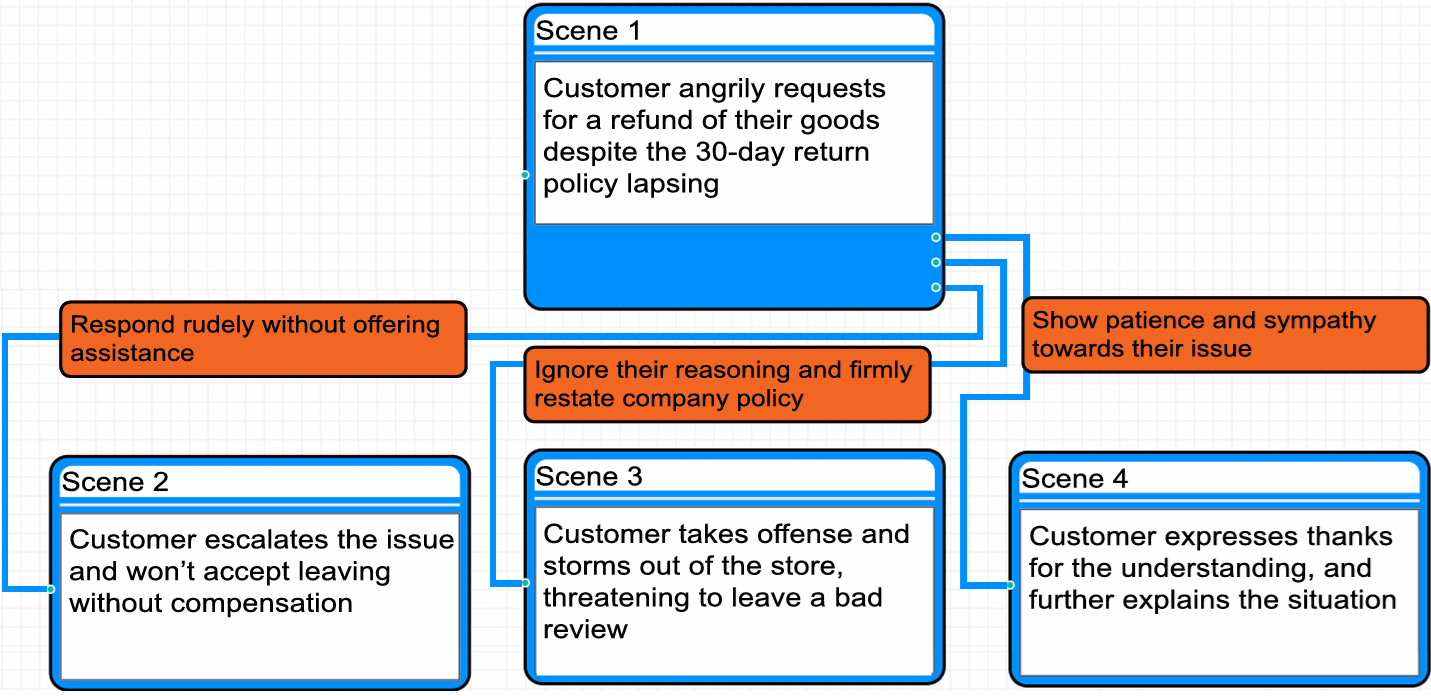}
    \caption{Narrative graph for customer service example}
    \label{fig:narrativeGraph}
\end{figure}
\\
\noindent
\textbf{Narrative graph:} The narrative graph (Figure~\ref{fig:narrativeGraph}) serves as the visual interface of the front-end builder, providing a graphical representation of all possible interactions within a scenario. Each node of the narrative graph represents a specific scenario, and the edges define the transition to the successor scenario based on the user's (student's) response.   For example, when handling an angry customer, the student might be patient, respond rudely, or ignore the complaint, leading to different successor scenarios within the narrative graph. 
% \MET{Does the user get to select from multiple responses? Or can they type in anything and then we analyze the free text? Oh - this is clarified below. Maybe clarify in this paragraph too?} 
The graph illustrates these options along with the corresponding reactions from the customer for each choice. Narrative graphs can be as detailed and complex as needed, allowing instructors to model a wide range of interactions and responses based on the training objectives. Instructors can add, delete, or modify nodes and edges in the narrative graph to tailor a scenario to be simple, complex, or appropriately aligned with a student's skill level. 
% Additional context related to the narrative, such as the scenario description and client persona can be defined to further influence the responses generated by the LLM.<-- this is in the front-end builder.
% %%% Chris stopped here -----
% Initializing the narrative Builder: initializing the graph template with a specific scenario, builds from scratch by instructor or generated by LLM using a prompt for a specific scenario, graph refinement by human to better cater to the scenario\\
%User-interface and feedback: Feedback about the student's 

\begin{figure}[tbp]
    \centering
    \includegraphics[width=.75\linewidth]{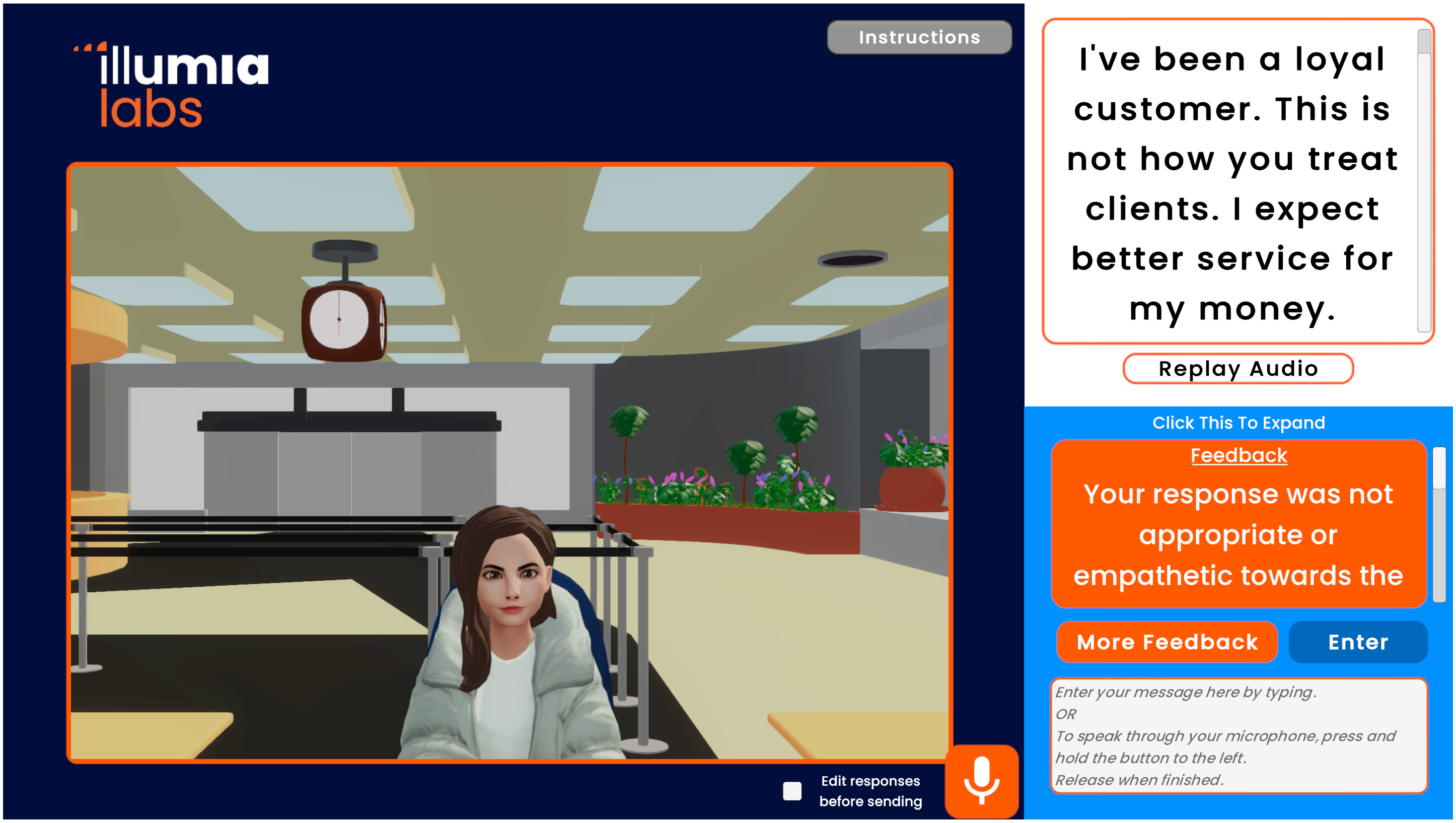}
    \caption{User-interface for GLOSS}
    \label{fig:userinterface}
\end{figure}

\noindent \textbf{Conversational simulator and feedback}: The conversational simulator (Figure~\ref{fig:userinterface}) is an interface for students to practice specific training scenarios from the built-in narrative graph. The student can enter their response through a text box or a microphone. Using the front-end builder, the instructor defines whether the dialogue should be strict and constrained, requiring precise phrasing, or more flexible, allowing the avatar (powered by the LLM) to generate dynamic, novel dialogues on the fly by mapping specific paths within the narrative graph. Furthermore, the system supports the automatic, dynamic generation of new paths in the narrative graph when existing options do not align with the user's response. These new transitions are added to the narrative graph by the LLM based on the intent of the user's response.
% \MET{Is this automatic? Can we say sometihng about how we determine whether an answer fits into one of the existing cases, or is different enough that we need to generate a new case?}
An avatar within the user interface reacts appropriately, simulating a real-world scenario. Feedback on the student's response is provided through an LLM using an independent prompt, allowing the student to adjust their response based on the feedback received. The interface enables users to practice uncomfortable, emotionally charged situations or culturally sensitive scenarios to approach through regular role-playing, offering a safe training environment. These simulations can be accessed from a desktop on a web browser via WebGL, or through immersive VR experiences.
\\
\noindent \textbf{Analysis tool}: The analysis tool visualizes a user's training conversation by illustrating their path through the narrative graph, serving as a mechanism to provide delayed feedback~\cite{hattie:07}.
% Figure xxx depicts an example conversation where the user diverts from the original graph, and the system generates new nodes to deliver the scenario. (NOTE: WILL MAKE A NEW GRAPH TO FIT THIS SENTENCE, IF DESIRED).
Instructors can assess student responses and determine if the content generated by the system reflect training goals. Instructor-in-the-loop workflows can incorporate this feedback loop to iteratively improve the narrative graph. The analysis tool also enables students to reflect on their conversations with the simulator, allowing them to enhance their social skills and improve their performance during a post-training review.

% \begin{figure}
%     \centering
%     \includegraphics[width=1\linewidth]{CameraReady/LaTeX/img/narrative_graph_analysis.PNG}
%     \caption{analysis tool}
%     \label{fig:analysisTool}
% \end{figure}
% (MG) not sure if we want this figure, if so will refactor it to properly match the narrative graph used in fig 2

% Feedback about the student's response is provided using an LLM (using an independent prompt).  The dialog can either be as strict as the instructor wants or as general as fitting a template but letting the LLM generate dialogues on the fly
% screen capture of the user interface (Avatar)

\section{Conclusion and Future Work}

The GLOSS framework leverages LLMs to create an instructor-in-the-loop tutoring system aimed at training social skills. GLOSS enables instructors to dynamically design realistic scenarios for students to navigate, facilitating rehearsal, providing immediate feedback and evaluation, and visualizing performance for both students and instructors. Future work includes enhancing the avatar's responses to student interactions and finding an optimal balance between fully scripted and open interactions. We also plan to refine immediate and delayed feedback mechanisms and conduct user studies to assess effectiveness for students and instructors~\cite{paramythis:10}.

% \textcolor{red}{My two main areas for future work would be 1. improving generated responses, through some more sophisticated system supplementing the LLM, and 2. further instructor-in-the-loop ideas such as the design of the front-end application from a co-creative tool and HCI perspective, as well as further work we can do with the feedback loop}

% \MET{What about a user study to evaluate the overall performance of the system, whether instructors can use it effectively, and/or if it helps students learn? Could cite: Paramythis+, 2010.}
%Paramythis, A., Weibelzahl, S. & Masthoff, J. Layered evaluation of interactive adaptive systems: framework and formative methods. User Model User-Adap Inter 20, 383–453 (2010). https://doi.org/10.1007/s11257-010-9082-4 }

\section{Acknowledgements}
This work has been completed for Illumia Labs Inc. as part of their advanced training platform. Part of this work has taken place in both the Intelligent Robot Learning (IRL) and EdTeKLA Lab at the University of Alberta, which are supported in part by grants from Amii; a Canada CIFAR AI Chair, Amii; DRAC; Mitacs; and NSERC. 

\bibliography{aaai25}

\begin{thebibliography}{13}
\providecommand{\natexlab}[1]{#1}

\bibitem[{{Battelle}(2019)}]{partner:19}
{Battelle}. 2019.
\newblock Framework for 21st Century Learning.
\newblock \url{https://www.battelleforkids.org/wp-content/uploads/2023/11/P21_Framework_Brief.pdf}.
\newblock Accessed: 2024-09-28.

\bibitem[{Georgila et~al.(2019)Georgila, Core, Nye, Karumbaiah, Auerbach, and Ram}]{georgila:19}
Georgila, K.; Core, M.~G.; Nye, B.~D.; Karumbaiah, S.; Auerbach, D.; and Ram, M. 2019.
\newblock Using reinforcement learning to optimize the policies of an intelligent tutoring system for interpersonal skills training.
\newblock In \emph{Proceedings of the 18th International Conference on Autonomous Agents and MultiAgent Systems}, 737--745.

\bibitem[{Hattie and Timperley(2007)}]{hattie:07}
Hattie, J.; and Timperley, H. 2007.
\newblock The power of feedback.
\newblock \emph{Review of educational research}, 77(1): 81--112.

\bibitem[{Herrity(2024)}]{indeed:24}
Herrity, J. 2024.
\newblock What are Social Skills? Definition and Examples.
\newblock \url{https://shorturl.at/mMTvm}.
\newblock Accessed: 2024-10-02.

\bibitem[{Lane et~al.(2007)Lane, Core, Gomboc, Karnavat, and Rosenberg}]{lane:07}
Lane, H.~C.; Core, M.~G.; Gomboc, D.; Karnavat, A.; and Rosenberg, M. 2007.
\newblock Intelligent tutoring for interpersonal and intercultural skills.
\newblock In \emph{Interservice/Industry Training, Simulationand Education Conference (I/ITSEC)}, volume 111.

\bibitem[{Leandro et~al.(2024)Leandro, Rao, Xu, Xu, Jojic, Brockett, and Dolan}]{Leandro:24}
Leandro, J.; Rao, S.; Xu, M.; Xu, W.; Jojic, N.; Brockett, C.; and Dolan, B. 2024.
\newblock GENEVA: GENErating and Visualizing branching narratives using LLMs.
\newblock In \emph{2024 IEEE Conference on Games (CoG)}, 1--5.

\bibitem[{Paramythis, Weibelzahl, and Masthoff(2010)}]{paramythis:10}
Paramythis, A.; Weibelzahl, S.; and Masthoff, J. 2010.
\newblock Layered evaluation of interactive adaptive systems: framework and formative methods.
\newblock \emph{User Modeling and User-Adapted Interaction}, 20: 383--453.

\bibitem[{Park et~al.(2023)Park, O'Brien, Cai, Morris, Liang, and Bernstein}]{Park:23}
Park, J.~S.; O'Brien, J.; Cai, C.~J.; Morris, M.~R.; Liang, P.; and Bernstein, M.~S. 2023.
\newblock Generative Agents: Interactive Simulacra of Human Behavior.
\newblock In \emph{Proceedings of the 36th Annual ACM Symposium on User Interface Software and Technology}, UIST '23. New York, NY, USA: Association for Computing Machinery.
\newblock ISBN 9798400701320.

\bibitem[{Sanchez et~al.(2014)Sanchez, Bartel, Brown, and DeRosier}]{sanchez:14}
Sanchez, R.~P.; Bartel, C.~M.; Brown, E.; and DeRosier, M. 2014.
\newblock The acceptability and efficacy of an intelligent social tutoring system.
\newblock \emph{Computers \& Education}, 78: 321--332.

\bibitem[{Sun, Zhan, and Such(2024)}]{Sun:24}
Sun, G.; Zhan, X.; and Such, J. 2024.
\newblock Building Better AI Agents: A Provocation on the Utilisation of Persona in LLM-based Conversational Agents.
\newblock In \emph{Proceedings of the 6th ACM Conference on Conversational User Interfaces}, CUI '24. New York, NY, USA: Association for Computing Machinery.
\newblock ISBN 9798400705113.

\bibitem[{Tanaka et~al.(2023)Tanaka, Saga, Iwauchi, Honda, Morimoto, Matsuda, Uratani, Okazaki, Nakamura et~al.}]{tanaka:23}
Tanaka, H.; Saga, T.; Iwauchi, K.; Honda, M.; Morimoto, T.; Matsuda, Y.; Uratani, M.; Okazaki, K.; Nakamura, S.; et~al. 2023.
\newblock The validation of automated social skills training in members of the general population over 4 weeks: Comparative study.
\newblock \emph{JMIR Formative Research}, 7(1): e44857.

\bibitem[{Tenhula and Bellack(2008)}]{tenhula:08}
Tenhula, W.~N.; and Bellack, A.~S. 2008.
\newblock Social skills training.
\newblock \emph{Clinical handbook of schizophrenia}, 240.

\bibitem[{Thornhill-Miller et~al.(2023)Thornhill-Miller, Camarda, Mercier, Burkhardt, Morisseau, Bourgeois-Bougrine, Vinchon, El~Hayek, Augereau-Landais, Mourey et~al.}]{thornhill:23}
Thornhill-Miller, B.; Camarda, A.; Mercier, M.; Burkhardt, J.-M.; Morisseau, T.; Bourgeois-Bougrine, S.; Vinchon, F.; El~Hayek, S.; Augereau-Landais, M.; Mourey, F.; et~al. 2023.
\newblock Creativity, critical thinking, communication, and collaboration: assessment, certification, and promotion of 21st century skills for the future of work and education.
\newblock \emph{Journal of Intelligence}, 11(3): 54.

\end{thebibliography}

\end{document}